\documentclass[10pt, a4paper]{article}
\usepackage{stylesheet/lrec}
\usepackage{multibib}
\newcites{languageresource}{Language Resources}
\usepackage{graphicx}
\usepackage{tabularx}
\usepackage{soul}

\usepackage{epstopdf}
\usepackage[latin1]{inputenc}

\usepackage{hyperref}
\usepackage{xstring}

\usepackage{booktabs}
\usepackage{multirow}
\usepackage{amsmath}
\usepackage{color}
\usepackage{makecell}
\usepackage{mathtools}

\hypersetup{draft} 

\newcommand{\mapping}{{\sc EmoMap}}

\newcommand{\isr}{ISR$_{\text{min}}$}

\title{Representation Mapping: \\ A Novel Approach to Generate High-Quality Multi-Lingual Emotion Lexicons}

\name{Sven Buechel \& Udo Hahn}

\address{
{\tt\{sven.buechel|udo.hahn\}@uni-jena.de}\\
Jena University Language and Information Engineering (JULIE) Lab\\
F\"urstengraben 27, 07743 Jena, Germany\\
\texttt{http://www.julielab.de}
}

\abstract{
In the past years, sentiment analysis has increasingly shifted attention to representational frameworks more expressive than semantic polarity (being positive, negative or neutral). However, these richer formats (like Basic Emotions or Valence-Arousal-Dominance, and variants therefrom), rooted in psychological research, tend to proliferate the number of representation schemes for emotion encoding.
Thus, a large amount of representationally incompatible emotion lexicons has been developed by various research groups adopting one or the other emotion representation format. As a consequence, the reusability of these resources decreases as does the comparability of systems using them.
In this paper, we propose to solve this dilemma by methods and tools which map different representation formats onto each other for the sake of mutual compatibility and interoperability of language resources.
We present the first large-scale investigation of such representation mappings for four typologically diverse languages and find evidence that our approach produces (near-)gold quality emotion lexicons, even in crosslingual settings. Finally, we use our models to create new lexicons for eight typologically diverse languages.
\\ \newline \Keywords{Automatic Construction of Emotion Lexicons, Representation Mapping, Models of Emotion} }

\begin{document}

\maketitleabstract

\section{Introduction}
\label{sec:intro}

In the past two decades, the NLP-based analysis and prediction of affective states, as performed by sentiment analysis systems, has received enormous interest \cite{Liu15}. Starting with simple positive-negative polarity distinctions on the word or text level \cite{Hatzivassiloglou97,Pang02}, research in sentiment analysis has since then shifted towards more nuanced and challenging tasks, e.g., sentiment compositionality \cite{Socher13}, aspect-level assessments \cite{Schouten16} or stance detection \cite{Sobhani16}. In parallel, psychologically more advanced and more expressive  {\it representation formats} for affective states have been proposed, like Basic Emotions \cite{Ekman92} or Valence-Arousal-Dominance \cite{Bradley94}. 
However, there is currently no consensus in the literature what scheme should be used as a common ground. Rather, there are a multitude of competing formats often motivated by the needs of concrete applications or the availability of user-labeled social media data \cite{Desmet13,Li16}.

While such decisions for a specific format may be perfectly reasonable in a specific research setting, on the flip-side, this proliferation of competing formats may seriously hamper progress in sentiment analysis for two reasons, at least. 
First, language resources are less reusable (if at all) as gold standards and, second, with the growing number of representation formats meaningful comparisons between predictive systems become harder (if not impossible).

One way to resolve this dilemma is to develop techniques to automatically translate between such formats. This task of {\it emotion representation mapping} (\mapping{}) was introduced only very recently to NLP by \newcite{Buechel17eacl}. Their work came up with an emotion-labeled corpus which, in part, is annotated with two different emotion formats both being highly predictive for each other. In a follow-up study, \newcite{Buechel17cogsci} examined the potential of \mapping{} as a substitute for manual annotation, yet their comparison was restricted to only two emotion lexicons.
Comparable work has (to the best of our knowledge) only been done in psychology. However, this stream of work does not target the goal of predictive modeling \cite{Stevenson07,Pinheiro17}. 
In NLP, a task related to \mapping{} is {\it emotion prediction} on the level of words, sentences, or texts \cite{Wang16words,Sedoc17} where, in contrast to \mapping{}, the target unit does not already need to bear annotations from another format. Thus, emotion prediction algorithms constitute a reasonable baseline for \mapping{} (see Section \ref{sec:experiments}).

This contribution puts emphasis on emotion lexicons developed in psychology. Although highly relevant for sentiment analysis, those resources have mostly been neglected by NLP researchers as the discussion of related work in Section \ref{sec:data} reveals.
Making use of this valuable work, we here conduct the first thorough evaluation of \mapping{} for emotion lexicon construction on four typologically diverse languages and find strong evidence that the quality of the output we generate is on a par with a gold standard when compared to human performance (see Section \ref{sec:experiments}).
Finally, we exploit our models to create novel emotion lexicons for eight different languages (including low-resource ones; Section \ref{sec:resource_construction}). The lexicons as well as the source code for building them are publicly available (see Section \ref{sec:conclusion}).

\section{Data}
\label{sec:data}

\begin{figure}
	\begin{center}
		\includegraphics[width=.45\textwidth]{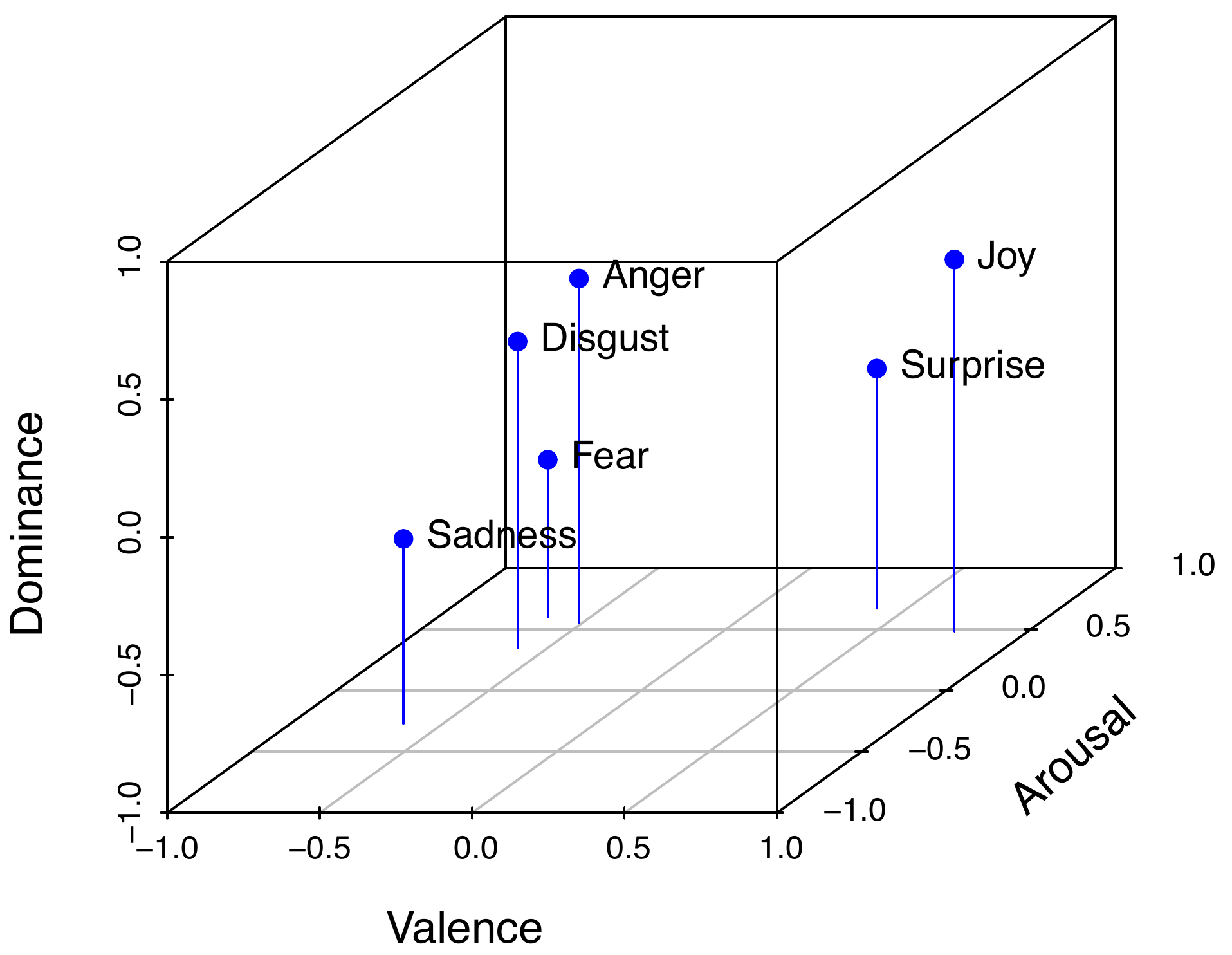}
				\vspace{-9pt}
		\caption{ Affective space spanned by the Valence-Arousal-Dominance model, together with the position of six Basic Emotions;  
			positions determined by \protect\newcite{Russell77}; 
			figure adapted from \protect\newcite{Buechel16ecai}. 
			\label{fig:vadcube}}
		\vspace{-19pt}
	\end{center}
\end{figure}

Models of emotion are typically subdivided into {\it discrete} (or {\it categorical}) and {\it dimensional} ones \cite{Stevenson07,Calvo13}. Discrete models are centered around particular sets of emotional categories deemed fundamental. \newcite{Ekman92}, for instance, identifies six {\it Basic Emotions} (Joy, Anger, Sadness, Fear, Disgust and Surprise). 
In contrast, dimensional models consider emotions to be composed out of several influencing factors (mainly two or three). These are often referred to as {\it Valence} (corresponding to the concept of polarity), {\it Arousal} (a calm--excited scale), and {\it Dominance} (perceived degree of control over a (social) situation)---the VAD model (see Figure \ref{fig:vadcube} for an illustration of the relationship between VAD dimensions and Basic Emotion categories).
The last dimension, Dominance, is sometimes omitted, leading to the VA model.

In contrast to NLP where many different formats are being used lexical resources in psychology almost exclusively subscribe to VA(D) or Basic Emotions (typically omitting Surprise; the BE5 format). 
Over the years, a considerable number of resources built on these premises have emerged from psychological research labs for various languages. 
Table \ref{tab:lexicons} enumerates published resources based on these two approaches (27 in total covering 13 languages, including low-resource ones such as Finnish and Indonesian).
To the best of our knowledge, the vast majority of them has neither been used nor referenced in NLP research.

\begin{table}
 \centering
 	\begin{tabular}{|p{3.8cm}p{.5cm}p{.9cm}r|}
 	\hline
 	 Reference &	Lang.\ & Format & ~~\# Entries \\
 	\hline \hline
 	\newcite{Warriner13}	& en & VAD & 13,915 \\
 	\newcite{Stevenson07} 	& en & BE5 &  1,034 \\
 	\newcite{Bradley99anew} & en & VAD &  1,034 \\
 	\newcite{Stadthagen16} 	& es & VA  & 14,031 \\
 	\newcite{Ferre16}  		& es & BE5 &  2,266 \\
 	\newcite{Guasch15} 		& es & VA  &  1,400\\
 	\newcite{Redondo07} 	& es & VAD &  1,034 \\
 	\newcite{Hinojosa16} 	& es & VA+BE5 & 875\\
 	\newcite{Hinojosa16dom} & es & \hspace*{13.1pt}+D & 875\\
 	\newcite{Vo09} 				& de & VA  & 2,902 \\
 	\newcite{Briesemeister11}  	& de & BE5 & 1,958 \\
 	\newcite{Schmidtke14}  		& de & VAD & 1,003 \\
 	\newcite{Kanske10}  		& de & VA  & 1,000 \\
 	\newcite{Imbir16} 		& pl & VAD & 4,905 \\
 	\newcite{Riegel15} 		& pl & VA  & 2,902 \\
 	\newcite{Wierzba15}		& pl & BE5 & 2,902 \\
 	\newcite{Yu16}			& zh & VA & 2,802\\
 	\newcite{Yao16} 		& zh & VA  & 1,100 \\
 	\newcite{Monnier14} & fr & VA & 1,031  \\
	\newcite{Ric13} & fr & V+BE5 & 524  \\
 	\newcite{Moors13} 		& nl & VAD & 4,299\\
 	\newcite{Sianipar16}	& id & VAD & 1,490 \\ 
 	\newcite{Palogiannidi16} & gr & VAD & 1,034  \\
 	\newcite{Montefinese14}	& it & VAD & 1,121\\
	\newcite{Soares12}		& pt & VAD & 1,034\\ 
 	\newcite{Eilola10} 		& fi & VA  &   210\\
 	\newcite{Davidson14}	& sv & VA  &   100\\
 	\hline
 	\end{tabular}
 	\caption{\label{tab:lexicons}
 	List of VA(D) and BE5 lexicons with empirically gathered ratings from human subjects; including  reference, language code (according to ISO 639-1), emotion representation format, and number of lexical entries.
 	}
 \end{table}

In this paper, we restrict ourselves to the VAD and BE5 format. In more detail (following the conventions of our emotion lexicons), each VAD dimension receives a value from the interval $[1,9]$ where `1' means ``most negative/calm/submissive'', `9' means ``most positive/excited/dominant'' and `5' means ``neutral''. Conversely, values for  BE5 categories range in the interval $[1,5]$ where `1' means ``absence'' and `5' means ``most extreme'' expression of the respective emotion.\footnote{
Although these intervals are fairly well established conventions, in some data sets different rating scales are used, nevertheless. In these cases, we linearly transformed the ratings so that they match the defined intervals.} 
Consequently, the VAD and BE5 formats are conceptually different from one another insofar as VAD dimensions are bi-polar, whereas BE5 categories are uni-polar.

Our work is based on the condition that some pairs of data sets in Table \ref{tab:lexicons} are complementary in the sense that, when combining these lexicons, a subset of the entries they contain are then described according to {\it both} emotion formats, VAD {\it and} BE5.
This condition is illustrated for three lexical items in Table \ref{tab:examples}.

\begin{table}[h!]
	\small
	\centering
	\begin{tabular}{|l|ccc|ccccc|}
		\hline 
		Word & V & A & D & J & A & S & F & D\\
		\hline \hline
		\textit{sunshine} 	& 8.1 & 5.3 & 5.4 & 4.2 & 1.2 & 1.3 & 1.3 & 1.2\\
		\textit{terrorism}	& 1.6 & 7.4 & 2.7 & 1.1 & 3.1 & 3.4 & 3.7 & 2.7\\
		\textit{orgasm} 	& 8.0 & 7.2 & 5.8 & 4.2 & 1.3 & 1.3 & 1.5 & 1.2\\
		\hline
	\end{tabular}
	\caption{Three lexical items and their emotion values in VAD (second column group) and BE5 (third column group) format. VAD scores are taken from \protect\newcite{Warriner13}, BE5 scores were automatically derived (see Section \ref{sec:resource_construction}).
		\label{tab:examples}}
	\vspace{-8pt}
\end{table}

From the resources listed in Table \ref{tab:lexicons}, we identified such complementary pairs and merged them into four lexicons which serve as gold data for the subsequent experiments: 

\begin{itemize}
\item \textbf{English}: \newcite{Bradley99anew} intersected with \newcite{Stevenson07} yielded 1,034 overlapping entries.
\vspace*{-6pt}
\item \textbf{Spanish}: \newcite{Redondo07} intersected with \newcite{Ferre16} yielded 1,012 overlapping entries.
\vspace*{-6pt}
\item \textbf{Polish}: \newcite{Imbir16} intersected with \newcite{Wierzba15} yielded 1,272 overlapping entries.
\vspace*{-6pt}
\item \textbf{German}: \newcite{Schmidtke14} intersected with \newcite{Briesemeister11} yielded 318 overlapping entries.
\end{itemize}

\begin{figure}[]
	\begin{center}
	\fbox{
	\includegraphics[width=.45\textwidth]{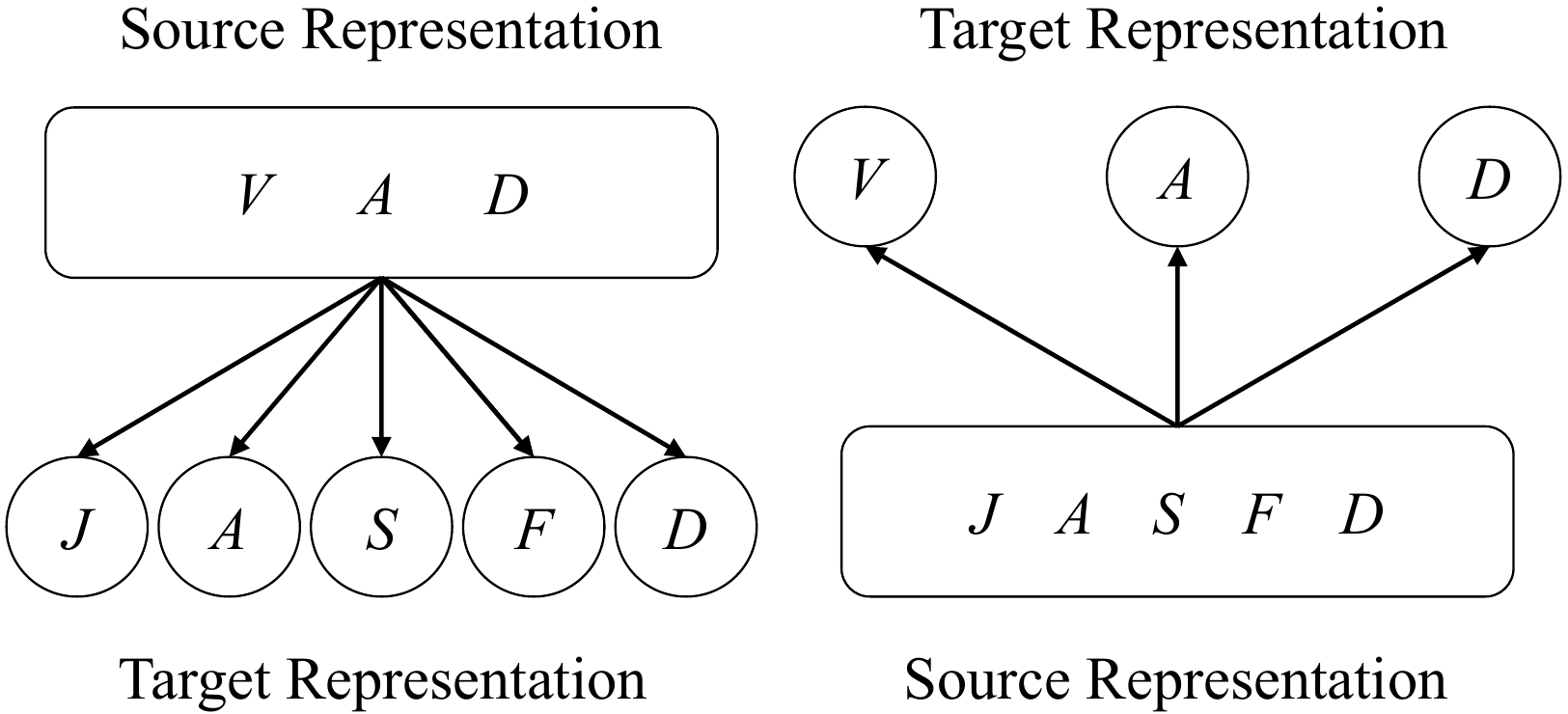}
	}
	\caption{\label{fig:mapping_scheme}
	Illustration of the two representation mapping procedures: VAD2BE5 (left) vs.\ BE52VAD (right). Each arrow represents an individual kNN model. 
	}
	\vspace*{-15pt}

	\end{center}
\end{figure}

\section{Method}
\label{sec:method}

Given an emotion lexicon in VAD format, our goal is to map its ratings onto the BE5 format and vice versa. We employ a simple, yet surprisingly efficient, method proposed by \newcite{Buechel17eacl}: For each of the dimensions or categories of the target representation (VAD or BE5, respectively), we train a single supervised model which employs each of the dimensions/categories of the source representation as features (e.g., {\it one} model to predict Joy, given Valence, Arousal, and Dominance scores as input; see Figure \ref{fig:mapping_scheme} for a graphical illustration of the general scheme).

In a pilot study, we compared different learning algorithms including linear regression, \textit{k} nearest neighbor regression (kNN), support vector regression (using different kernels), random forests, as well as feed-forward neural networks. 
To our surprise, all of them performed equally well (with only negligible differences). Thus, kNN was selected due to its simplicity.\footnote{We use the \texttt{scikit-learn.org} implementation.}
Note that the feature set is extremely small (either three or five variables for mapping {\it onto} BE5 or VAD, respectively) so that using more complex methods (e.g., more sophisticated neural architectures) seems a waste of efforts.

\begin{table*}
\begin{centering}
\begin{tabular}{|lrrrr|}
\hline
{} &  Val &  Aro &  Dom &    \#Overlap \\
\hline \hline
\newcite{Imbir16} vs. \newcite{Riegel15}        &     {\bf .948} &     .733 &       ---&  1,272 \\
\newcite{Guasch15} vs. \newcite{Stadthagen16}   &     .949 &     {\bf .875} &       ---&  1,298 \\
\newcite{Bradley99anew} vs. \newcite{Warriner13}       &     .952 &     .760 &       {\bf .794} &  1,027 \\
\newcite{Guasch15} vs. \newcite{Hinojosa16}     &     .968 &     .777 &       ---&   134 \\
\newcite{Guasch15} vs. \newcite{Redondo07}      &     .969 &     .844 &       ---&   316 \\
\newcite{Hinojosa16} vs. \newcite{Stadthagen16} &     .970 &     {\bf .709} &       ---&   636 \\
\newcite{Schmidtke14} vs. \newcite{Kanske10}        &     .971 &     .788 &       ---&   169 \\
\newcite{Redondo07} vs. \newcite{Stadthagen16}  &     {\bf .976} &     .755 &       ---&  1,010 \\
\hline
\end{tabular}
\caption{\label{tab:isr}
Inter-study reliabilities between different data sets (measured in $r$). Minimum and maximum values for each VAD dimension (respectively) in bold. }
\end{centering}
\end{table*}

\begin{figure*}
\begin{center}
\fbox{
\includegraphics[width=.97\textwidth]{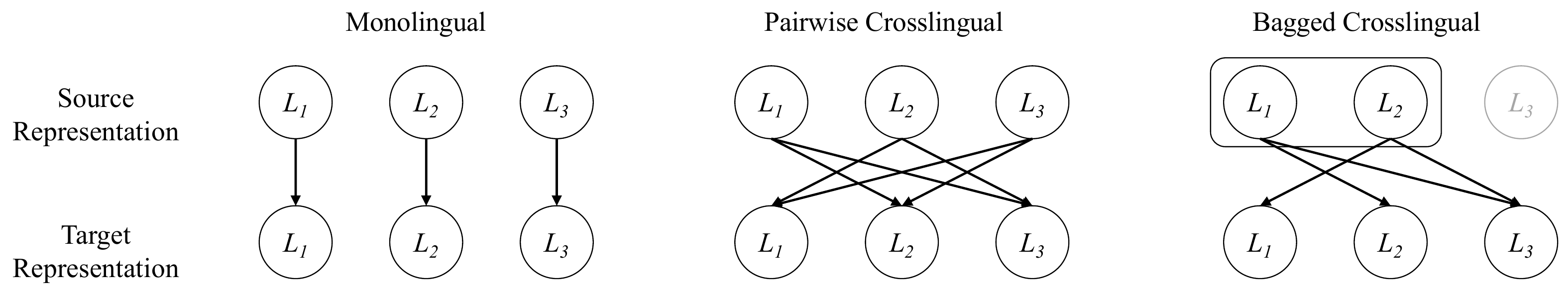}
\vspace*{-15pt}
}
\caption{\label{fig:strategies}
Illustration of the three mapping strategies applied in Section \ref{sec:experiments} exemplified for the languages $L_1$, $L_2$ and $L_3$.}
\vspace*{-15pt}
\end{center}
\end{figure*}

\begin{table*}
\centering
\begin{tabular}{|p{2cm}l|llll|llllll|}
\hline
 Experiment & Language &    Val &   Aro &  Dom &    {\bf Av$_{\text{VAD}}$} &        Joy &      Anger &    Sadn&       Fear &    Disg&   {\bf  Av$_{\text{BE5}}$} \\
\hline\hline
\multirow{4}{*}{monolingual} 	&English &    .966* &    .723 &      .833*&    {\bf.841} & .958 &  .870 &    .864 & .864 &    .790 &    {\bf.869} \\
&Spanish &    .970* &    .736 &      .855* &    {\bf.854} & .957 &  .847 &    .828 & .870 &    .744 &    {\bf.849} \\
&Polish  &    .944 &    .761* &      .740 &    {\bf.815} & .932 &  .845 &    .803 & .784 &    .814 &    {\bf.836} \\
&German  &    .950 &    .762* &      .637 &    {\bf.783} & .923 &  .793 &    .680 & .851 &    .602 &    {\bf.770} \\
\hline
\multirow{12}{*}{\makecell{crosslingual\\ (pairwise)}}	
&es2en &     .963* &     .714 &       .794 &     {\bf.824} & .948 &   .830 &     .853 &  .835 &     .780 &     {\bf.849} \\	
&pl2en &     .962* &     .598 &       .776 &     {\bf.778} & .955 &   .845 &     .836 &  .832 &     .765 &     {\bf.847} \\
&de2en &     .952 &     .445 &       .762 &     {\bf.720} & .952 &   .861 &     .836 &  .855 &     .746 &     {\bf.850} \\
&en2es &     .966* &     .737* &       .811 &     {\bf.838} & .948 &   .791 &     .806 &  .826 &     .694 &     {\bf.813} \\
&pl2es &     .961* &     .634 &       .701 &     {\bf.765} & .941 &   .744 &     .763 &  .766 &     .665 &     {\bf.776} \\
&de2es &     .959* &     .498 &       .842* &     {\bf.767} & .942 &   .794 &     .785 &  .839 &     .640 &     {\bf.800} \\
&en2pl &     .938 &     .655 &       .653 &     {\bf.749} & .924 &   .816 &     .800 &  .751 &     .795 &     {\bf.817} \\
&es2pl &     .934 &     .663 &       .552 &     {\bf.717} & .918 &   .755 &     .762 &  .653 &     .768 &     {\bf.771} \\
&de2pl &     .920 &     .674 &       .497 &     {\bf.697} & .914 &   .815 &     .759 &  .700 &     .739 &     {\bf.785} \\
&en2de &     .940 &     .615 &       .583 &     {\bf.713} & .915 &   .789 &     .678 &  .849 &     .584 &     {\bf.763} \\
&es2de &     .953 &     .618 &       .645 &     {\bf.739} & .904 &   .789 &     .692 &  .840 &     .579 &     {\bf.761} \\
&pl2de &     .934 &     .691 &       .358 &     {\bf.661} & .907 &   .768 &     .655 &  .788 &     .529 &     {\bf.730} \\
\hline
\multirow{4}{*}{\makecell{crosslingual\\ (bagged)}}	&English &      .963* &      .714 &      .794 &      {\bf .824} &      .948 &      .830 &      .853 &      .835 &      .780 &      {\bf .849} \\
								&Spanish &      .966* &      .737* &      .811 &      {\bf .838} &      .948 &      .791 &      .806 &      .826 &      .694 &      {\bf .813} \\
								&Polish  &      .939 &      .645 &      .629 &      {\bf .738} &      .926 &      .781 &      .780 &      .700 &      .769 &      {\bf .791} \\
								&German  &      .949 &      .635 &      .632 &      {\bf .739} &      .917 &      .799 &      .692 &      .844 &      .551 &      {\bf .761} \\
\hline
\end{tabular}

\caption{Results of the monolingual (Section \ref{sec:exp.mono}) and crosslingual (Sections \ref{sec:exp.pairwise} and \ref{sec:exp.bagged}) evaluation in Pearson's $r$. Language `a2b' denotes mapping from language {\it a} (source) to language {\it b} (target). Significant values are marked with `*' (compared to \isr{}; $p < .05$; VAD only), averages over VAD and BE5 (respectively) in bold.\label{tab:results}}
\end{table*}

\section{Experiments}
\label{sec:experiments}

We here present the first large-scale evaluation of emotion representation mapping (\mapping). Our methodology, at the same time, leads to the automatic construction of emotion lexicons for four typologically diverse languages. We consider one monolingual and two crosslingual set-ups, i.e., training and testing data from the same or different language(s), respectively. Those three different mapping strategies are illustrated in Figure \ref{fig:strategies}.

The performance of the \mapping{} approach will be measured as Pearson correlation ($r$) between our automatically predicted values and human gold ratings. 
In general, the Pearson correlation between two data series $X=x_1, x_2, ..., x_n$ and $Y=y_1, y_2, ..., y_n$ takes values between $+1$ (perfect positive correlation) and $-1$ (perfect negative correlation). It is computed as 
\begin{equation}
r_{xy} \coloneqq \frac{\sum_{i=1}^n (x_i-\bar{x})(y_i-\bar{y})}{\sqrt{\sum_{i=1}^n(x_i-\bar{x})^2} \; \sqrt{\sum_{i=1}^n(y_i-\bar{y})^2}}
\end{equation}
where $\bar{x}$ and $\bar{y}$ denote the mean values for $X$ and $Y$, respectively.

These measurements will then be compared, first, with the current state-of-the-art in word-level emotion prediction (as baseline), and, second,  with human inter-study reliability (as ceiling). Both comparisons will, for different reasons, be limited to the VAD model.

\subsection{Baseline and Ceiling}
Word-level emotion prediction (automatically deriving the emotion of a word from scratch; see Section \ref{sec:intro}) serves as a reasonable baseline since it produces the same output as \mapping{}, yet does not require the target words to have already been annotated in a different emotion format (other than the output representation).

\newcite{Sedoc17} evaluated their approach to word-level emotion prediction on the data set compiled by \newcite{Bradley99anew} using 10-fold cross-validation. They  report measurements of $r=0.806$ for Valence and $r=0.615$ for Arousal. 
Concerning the other affective dimensions and categories, we are not aware of any other system predicting numerical scores for them. Thus, we will restrict our comparison to Valence and Arousal.

For comparison against the human ceiling, we found eight pairs of emotion lexicons with partially overlapping entries distributed over four languages. 
For each of these pairs, we computed their inter-study reliability (ISR), i.e., the Pearson correlation between the ratings from the two respective studies for each affective dimension (see Table \ref{tab:isr}).
Again, because we only found lexicons with overlapping VAD (not BE5) entries, we restrict this comparison to VAD representations.

We stipulate that for all ISR values from Table \ref{tab:isr}, the {\it minimum} for each affective dimension constitutes the most relevant score of comparison. The rationale for this assumption is as follows: If our approach happens to outperform this minimal value, one cannot be certain that manual annotation leads to better results than using our automatic procedure. In this situation, we assume that the computational approach would almost always be preferred over manual annotation efforts.
Accordingly, the following correlation values were identified as minimum inter-study reliabilities: $r=.948$ for Valence, $r=.709$ for Arousal, and $r=.794$ for Dominance (henceforth, jointly referred to as \isr).

Note that comparing against the ISR is a much harder test than comparing against inter-annotator agreement (IAA): Since the former is based on the mean rating of many raters, this aggregated judgment is more stable than individual ratings, thus resulting in higher correlation values compared to its IAA counterpart.\footnote{For numerical emotion ratings, IAA is typically computed in a leave-one-out fashion and can thus be interpreted as how well a single human annotator predicts the gold value \cite{Strapparava07,Buechel17eacl}.}

\subsection{Monolingual Evaluation}
\label{sec:exp.mono}

The first part of our analysis concerns train and test data originating from the same language (respectively data set); see left part of Figure \ref{fig:strategies}. For each of our (language-wise) four gold lexicons (cf.\ the final paragraph of Section  \ref{sec:data}), we train kNN models to map between VAD and BE5 representations back and forth according to the scheme from Figure \ref{fig:mapping_scheme}. 
Training and testing was done using 10-fold cross-validation (9:1 train/test split). The k-parameter was fixed to 20, based on a pilot study (eliminating the need for a dev set). The results are presented in Table \ref{tab:results}, upper section.

As can be seen, the outcome is overall favorable for our approach. In general, it works about equally well in both mapping directions (VAD2BE5 and BE52VAD) with average values (over VAD dimensions and BE5 categories, respectively) of $r \geq 77\%$. The results on the English and Spanish gold lexicons are better than for the Polish, yet worst for the German one (which is also the smallest).
In comparison with the baseline (see above; English data set only), our \mapping{}  approach performs more than 15 percentage points better for Valence and more than 10 percentage points better for Arousal.
Even more surprisingly, compared to the human ceiling, we find that our approach outperforms the \isr{} in 9 out of 12 cases (again, only failing to do so on the Polish and German data set).
In those 9 cases where we outperformed the \isr, we conducted a one-tailed  one sample t-test based on the 10 individual cross-validation results \cite{Dietterich98} finding significant differences in 6 of these cases ($p<.05$; marked with asterisk in Table \ref{tab:results}).

We conclude that, in the monolingual set-up, \mapping{} performs on a par with (if not superior to) manual annotation for mapping {\it onto} VAD. Thus, its results can be considered as true gold data. For BE5, we cannot draw the same conclusion due to a lack of data on inter-study reliability. However, since the performance figures for the VAD2BE5 mapping are equally high, we may quite safely assume that the Basic Emotion ratings can be attributed high quality as well.

\subsection{Pairwise Crosslingual Evaluation}
\label{sec:exp.pairwise}

In the two crosslingual  set-ups (training and test data drawn from different languages, respectively data sets), we make use of the fact that our models do not rely on any language-specific information since the categories/dimensions describe (supposedly universal) affective states rather than linguistic entities.
Thus, models trained on one language could, in theory, be applied to another without any adaptation. 

Let us, first, address {\it pairwise} comparisons. That is, for each language, we train our kNN models on the entirety of the respective data set and then test on all the remaining languages individually (illustrated in Figure \ref{fig:strategies}; resulting in a total of 12 language pairs). Since, this set-up uses fixed training and test sets, there is no need for cross-validation. The results are given in Table \ref{tab:results} (middle section).

Overall, the values remain astonishingly high. As can be seen, for mapping BE52VAD, the results are quite favorable for Valence with correlation values ranging well above $90\%$ of correlation. On this dimension, our approach still outperforms the baseline by over a $15\%$-points margin and even surpasses the human ceiling in more than half of the cases, five of them being statistically significant. Since different from Section \ref{sec:exp.mono}, we now have a fixed test set, we use a one-tailed z-test ($p<.05$) based on z-transformed correlation values \cite{Cohen95}.

In contrast to Valence, the performance for Arousal and Dominance may suffer quite substantially in the crosslingual approach, depending on the combination of training and testing languages. While there is almost no performance loss for combinations of English and Spanish, the correlation decreases the most for combinations of Polish and German (especially for predicting Dominance), possibly due to data sparsity of the lexicons involved.

This outcome led us to conclude that the relationship between VAD and BE5 ratings is not fully constant across different languages (respectively data sets). Rather it seems to depend on subtle semantic differences between the translational equivalents of the affective dimensions/categories, cultural differences, or variations in the annotation guidelines, suggesting that the above assumption of language independence (not so surprisingly) may not fully hold.

In contrast to these partly inconclusive results, the outcome for mapping VAD2BE5 is much more favorable for \mapping{} and easy to describe. Compared to the monolingual set-up (relative to the target language), the drop of the average performance amounts to only a few percentage points ($<5$ in most cases). Thus, the predictions for BE5 are much more robust compared to the VAD predictions which might be an effect of the respective source representation. 

We conclude that in the pairwise crosslingual set-up, \mapping{} still performs really well in many cases. Yet depending on the language pair, the performance may degrade (much more severely so for mapping BE52VAD).

\subsection{Bagged Crosslingual Evaluation}
\label{sec:exp.bagged}

As evident from the last section, the performance of our mapping approach may vary depending on source and target language. However, different from the last experiment, when constructing new emotion lexicons in a crosslingual fashion, there is no need to restrict the training set to only one language. Instead, because no language-specific features are used, we may merge training data from multiple languages if this leads to a more robust predictive model. However, since not all languages (respectively data sets) seem to match well, there is no guarantee that more data sets always help boosting performance.
In line with these considerations, the goal of the last experiment is to identify the best group of training data sets for automatically creating novel lexicons and to estimate their quality.

For each combination of gold lexicons of bag sizes two\footnote{If only one  lexicon would be used for training, this one could not also be used for testing, thus making the different combinations incomparable.} to four and each target language, we train our models on the entirety of the bag of training data (except the one designated for testing, should it also belong to the training data) and then test on the target language; see the third data scenario in the right part of Figure \ref{fig:strategies}.

In line with Section \ref{sec:exp.pairwise}, we found that the average performance over VAD behaved less robust across different combinations of training data (ranging between $r=.730$ to $.786$) compared to BE5 ($r=.771$ to $.806$).
Concerning the average over {\it all} emotions, the combination of Polish and German, once again, performed worst ($r=.755$), whereas the combination of English and Spanish worked best ($r=.794$; i.e., for testing on English, Spanish was used for training and vice versa, while for testing on the remaining languages, training was done on English {\it and} Spanish). Consequently, this combination of gold data was used for creating novel lexicons in the crosslingual set-up (see Section \ref{sec:resource_construction}).

Table \ref{tab:results} (bottom section) displays the results of this crosslingual experiment for the best performing bag of training data (comprising the English and Spanish gold lexicons, only). Thus, these performance data serve as an estimate of the quality of the novel  emotion lexicons presented in Section \ref{sec:resource_construction}
Overall, we find that the  results are again favorable for our approach. The correlation with the English and Spanish data set, not surprisingly, is stronger than with the Polish and German one, confirming that these data sets form a better basis for generalization (the effect being more obvious for VAD than for BE5). In general, Valence, Joy, and Anger can be predicted with consistently high correlation, whereas for the other dimensions/categories we find occasional negative outliers.

Comparing our results to human reliability (in VAD only), we find that our models are superior to human \isr{} in 7 from 12 cases (including all cases on the English and Spanish data set). In 3 of these cases, the difference is statistically significant ($p<.05$). 
In comparison to the baseline (on English, VA only), our approach  still clearly outperforms state-of-the-art word-level emotion prediction by a 15 and 10 percentage point margin for Valence and Arousal, respectively.

We conclude that even if no gold data for a given language are available, \mapping{} still performs comparably to human reliability when utilizing appropriate sets of training data. Some dimensions and categories seem to be reliable across data sets, whereas for others the performance may degrade, depending on the target data set. Yet, the lexicons derived in this set-up can still be attested  \textit{near-gold} quality.

\begin{table}[t]
\centering
	\begin{tabular}{|l|l|r|r|}
		\hline
		&Language & \#VAD & \#BE5\\
		\hline
		\hline
		\multirow{4}{*}{monolingual}	& English 	&  		& 12,888 	\\
									& Spanish 	& 1,254 & 			\\
									& German  	& 1,641 & 683		\\
									& Polish	& 		& 3,633		\\
		\hline
		\multirow{4}{*}{crosslingual}	& Italian 	&		& 1,121		\\
									& Portuguese&		& 1,034		\\
									& Dutch 	& 		& 4,299 	\\
									& Indonesian& 		& 1,490 	\\
		\hline
	\end{tabular}
	\caption{Automatically constructed gold quality lexicon resources, `\#' indicates the number of previously unrated lexical units for a specific representation format. \label{tab:resource_construction}}
\end{table}

\begin{table}
\begin{center}
\begin{tabular}{|lrrrrr|}
\hline
{} &   Joy &  Anger &  Sadness &  Fear &  Disgust \\
\hline\hline
Mean   &  2.11 &   1.62 &     1.61 &  1.66 &     1.59 \\
Median &  1.86 &   1.38 &     1.38 &  1.42 &     1.37 \\
Min    &  1.07 &   1.14 &     1.21 &  1.17 &     1.11 \\
Max    &  4.40 &   3.38 &     3.81 &  3.74 &     3.26 \\
StDev  &  0.79 &   0.47 &     0.47 &  0.49 &     0.47 \\
\hline
\end{tabular}
\caption{\label{tab:stats}
Descriptive statistics for the automatically constructed English BE5 lexicon.}
\end{center}
\end{table}

\begin{table}
\begin{center}
\small
\begin{tabular}{|l|l|l|l|l|}
\hline
       \hspace{-4pt}Joy\hspace{-4pt} &      \hspace{-4pt}Anger\hspace{-4pt} &      \hspace{-4pt}Sadness\hspace{-4pt} &          \hspace{-4pt}Fear\hspace{-4pt} &     \hspace{-4pt}Disgust\hspace{-4pt} \\
\hline\hline
 \hspace{-4pt}\textit{christmas}\hspace{-4pt} &     \hspace{-4pt}\textit{killer}\hspace{-4pt} &        \hspace{-4pt}\textit{chemo}\hspace{-4pt} &      \hspace{-4pt}\textit{insanity}\hspace{-4pt} &      \hspace{-4pt}\textit{felony}\hspace{-4pt} \\
 \hspace{-4pt}\textit{happiness}\hspace{-4pt} &       \hspace{-4pt}\textit{gang}\hspace{-4pt} &    \hspace{-4pt}\textit{worthless}\hspace{-4pt} &  \hspace{-4pt}\textit{motherfucker}\hspace{-4pt} &     \hspace{-4pt}\textit{enraged}\hspace{-4pt} \\
   \hspace{-4pt}\textit{magical}\hspace{-4pt} &    \hspace{-4pt}\textit{revenge}\hspace{-4pt} &    \hspace{-4pt}\textit{gonorrhea}\hspace{-4pt} &     \hspace{-4pt}\textit{terrorism}\hspace{-4pt} &  \hspace{-4pt}\textit{traitorous}\hspace{-4pt} \\
       \hspace{-4pt}\textit{fun}\hspace{-4pt} &        \hspace{-4pt}\textit{die}\hspace{-4pt} &       \hspace{-4pt}\textit{nausea}\hspace{-4pt} &      \hspace{-4pt}\textit{attacker}\hspace{-4pt} &  \hspace{-4pt}\textit{dishonesty}\hspace{-4pt} \\
 \hspace{-4pt}\textit{enjoyment}\hspace{-4pt} &   \hspace{-4pt}\textit{massacre}\hspace{-4pt} &        \hspace{-4pt}\textit{virus}\hspace{-4pt} &      \hspace{-4pt}\textit{bullshit}\hspace{-4pt} &  \hspace{-4pt}\textit{chauvinist}\hspace{-4pt} \\
     \hspace{-4pt}\textit{bonus}\hspace{-4pt} &   \hspace{-4pt}\textit{attacker}\hspace{-4pt} &   \hspace{-4pt}\textit{amputation}\hspace{-4pt} &     \hspace{-4pt}\textit{murderous}\hspace{-4pt} &    \hspace{-4pt}\textit{mistrust}\hspace{-4pt} \\
     \hspace{-4pt}\textit{oasis}\hspace{-4pt} &        \hspace{-4pt}\textit{sue}\hspace{-4pt} &  \hspace{-4pt}\textit{unhappiness}\hspace{-4pt} &     \hspace{-4pt}\textit{dangerous}\hspace{-4pt} &        \hspace{-4pt}\textit{gory}\hspace{-4pt} \\
 \hspace{-4pt}\textit{fantastic}\hspace{-4pt} &   \hspace{-4pt}\textit{hijacker}\hspace{-4pt} &   \hspace{-4pt}\textit{unsanitary}\hspace{-4pt} &       \hspace{-4pt}\textit{tragedy}\hspace{-4pt} &     \hspace{-4pt}\textit{hostile}\hspace{-4pt} \\
     \hspace{-4pt}\textit{happy}\hspace{-4pt} &     \hspace{-4pt}\textit{nigger}\hspace{-4pt} &     \hspace{-4pt}\textit{molester}\hspace{-4pt} &        \hspace{-4pt}\textit{arrest}\hspace{-4pt} &      \hspace{-4pt}\textit{racist}\hspace{-4pt} \\
  \hspace{-4pt}\textit{sunshine}\hspace{-4pt} &  \hspace{-4pt}\textit{penniless}\hspace{-4pt} &     \hspace{-4pt}\textit{lynching}\hspace{-4pt} &          \hspace{-4pt}\textit{rape}\hspace{-4pt} &   \hspace{-4pt}\textit{cellulite}\hspace{-4pt} \\
\hline

\end{tabular}
\caption{\label{tab:top_k}
Top 10 entries per Basic Emotion in automatically constructed English BE5 lexicon.}
\end{center}
\end{table}

\begin{table}
\begin{center}
\small
\begin{tabular}{|l|rrr|rrrrr|}
\hline
{} &    V &     A &     D &     J &     A &     S &     F &     D \\
\hline
V &  -  & --.18 & +.72 & +.92 & --.83 & --.82 & --.75 & --.87 \\
A &  -  &   -  & --.18 & --.03 & +.58 & +.46 & +.67 & +.41 \\
D &  -  &   -  &   -  & +.68 & --.66 & --.76 & --.68 & --.61 \\
\hline
J &  -  &   -  &   -  &   -  & --.66 & --.61 & --.59 & --.69 \\
A &  -  &   -  &   -  &   -  &   -  & +.92 & +.95 & +.91 \\
S &  -  &   -  &   -  &   -  &   -  &   -  & +.91 & +.85 \\
F &  -  &   -  &   -  &   -  &   -  &   -  &   -  & +.82 \\
D &  -  &   -  &   -  &   -  &   -  &   -  &   -  &   -  \\
\hline
\end{tabular}
\caption{\label{tab:correlation}
Correlation matrix (in $r$) for automatically constructed English BE5 (JASFD) lexicon combined with the data by \protect\newcite{Warriner13} (VAD).}
\vspace*{-15pt}
\end{center}
\end{table}

\section{ Construction of New Emotion Lexicons}
\label{sec:resource_construction}

After the positive evaluation of \mapping{} for four typologically diverse languages, our main contribution is to apply the created models to a wide variety of data sets which so far  bear emotion ratings for {\it one} format only (either VAD or BE5). Based on our preceding experiments, we claim that these have gold quality (using the monolingual approach, Section \ref{sec:exp.mono}) or near-gold quality (using the crosslingual approach, Section \ref{sec:exp.bagged}).
We constructed a total of nine emotion lexicons covering eight languages (including low-resource ones, such as Dutch and Indonesian). Table \ref{tab:resource_construction} depicts the number of lexical items for which we have generated previously unknown VAD or BE5 ratings per language. 
For illustration, we provide an analysis of the  English BE5 lexicon (by far the largest resource constructed in this manner) in the remainder of this section.

Table \ref{tab:stats} provides fundamental statistical characteristics of this newly developed data set. As can be seen, Joy ratings have higher mean, standard deviation and range than all the other categories. This suggests that a larger portion of lexical items expresses at least a moderate degree of Joy, whereas the other Basic Emotions are expressed less often and to a smaller extent. 
Table \ref{tab:top_k} lists the ten entries with the highest values for each Basic Emotion category. Obviously, the automatically derived ratings align well with our intuition, thus granting face validity to our approach.

Finally, Table \ref{tab:correlation} provides correlation values between the BE5 categories and VAD dimensions (ratings for the latter were taken from \newcite{Warriner13}). Joy displays a moderate negative correlation with the other Basic Emotions while these in turn have strong positive correlation among each other. Unsurprisingly, Valence displays a strong positive correlation with Joy and strong negative correlations with the remaining BE5 categories. Lastly, Arousal is uncorrelated with Joy but displays moderate positive correlation with Anger, Sadness, Fear and Disgust. These findings are consistent with empirically determined emotion values, thus validating our claims concerning the good quality of the constructed resources \cite{Wierzba15,Hinojosa16}.

\section{Conclusion}
\label{sec:conclusion}

Progress in emotion analysis is hampered by a multitude of heterogeneous and, in the end, mutually incompatible emotion representation formats. In this paper, we performed the first large-scale analysis of {\it representation mapping} as a means to mediate between these heterogeneous formats. Our simple, yet highly effective, supervised approach makes use of the wide range of emotion lexicons already developed in various psychology labs. 

We could show that, in the monolingual setup, automatic representation mapping outperforms human inter-study reliability and, therefore, produces \textit{gold quality} data. In the crosslingual set-up, our approach still performs comparable to manual annotation though less robust than in the first set-up for some affective dimensions or categories, thus rendering \textit{near-gold quality} entries. In both set-ups, mapping existing ratings to another format performs way better than the state-of-the-art in emotion prediction. Hence, we conjecture that our approach paves the way to greatly improve interoperability and re-use of lexical resources in this field.

Lastly, we applied our technique to produce (near) gold quality emotion lexicons for eight typologically diverse languages, including low-resourced ones. These resources (together with our code) are available via \texttt{github.com/JULIELab/EmoMap}.

\section{Bibliographical References}
\label{main:ref}

\bibliographystyle{stylesheet/lrec}
\bibliography{literatureSB-LREC18}

\end{document}